# DI-AA: An Interpretable White-box Attack for Fooling Deep Neural Networks

**Yixiang Wang,**[1] **Jiqiang Liu,**[1*] **Xiaolin Chang,**[1] **Jianhua Wang**, **Ricardo J. Rodríguez**[2]

[1]Beijing Key Laboratory of Security and Privacy in Intelligent Transportation, Beijing Jiaotong University, Beijing 100044, China

[2]Dept. of Computer Science and Systems Engineering, University of Zaragoza

[1]{18112047, jqliu, xlchang, 20112051}@bjtu.edu.cn, [2]rjrodriguez@unizar.es

**Abstract**

White-box Adversarial Example (AE) attacks towards Deep Neural Networks (DNNs) have a more powerful destructive capacity than black-box AE attacks in the fields of AE strategies. However, almost all the white-box approaches lack *interpretation from the point of view of DNNs*. That is, adversaries did not investigate the attacks from the perspective of interpretable features, and few of these approaches considered what features the DNN actually *learns*. In this paper, we propose an interpretable white-box AE attack approach, DI-AA, which explores the application of the interpretable approach of the deep Taylor decomposition in the selection of the most contributing features and adopts the Lagrangian relaxation optimization of the logit output and $L_p$ norm to further decrease the perturbation. We compare DI-AA with six baseline attacks (including the state-of-the-art attack AutoAttack) on three datasets. Experimental results reveal that our proposed approach can 1) attack non-robust models with comparatively low perturbation, where the perturbation is closer to or lower than the AutoAttack approach; 2) break the TRADES adversarial training models with the highest success rate; 3) the generated AE can reduce the robust accuracy of the robust black-box models by 16% to 31% in the black-box transfer attack.

## 1. Introduction

Recently, notable advances in the development of Deep Neural Networks (DNNs) have led to breakthroughs in various areas such as bioinformatics (Senior et al. 2020), language learning (Lei et al. 2021) and causal inference (Luo, Peng, and Ma 2020). However, the emergence of adversarial examples (AE) threatens the widespread deployment of security-sensitive DNN-based applications (Szegedy et al. 2014; Goodfellow, Shlens, and Szegedy 2015). In AE attacks, a clean input with a small, unnoticeable perturbation will mislead a well-trained DNN, and the vulnerability and counterintuitive behavior of DNNs towards AE make it difficult for users to trust DNN decisions.

Depending on the ability of the adversaries, there are two types of attack: white-box attack (Szegedy et al. 2014; Kurakin 2017; Carlini and Wagner 2017; Madry et al. 2018; Croce and Hein 2020b) and black-box attack (Chen et al. 2017; Moon, An, and Song 2019; Chen et al. 2020; Wu et al. 2020). The adversaries have complete knowledge of the target model and data information in white-box settings. In contrast, in black-box settings, adversaries can transfer the generated AE to the unknown deployed model based on AE transferability (Wu et al. 2020). Empirically, white-box attacks are more powerful to attack a robust model than black-box attacks (Croce and Hein 2020b).

In terms of white-box attacks, Croce and Hein 2020a further divide them into two classes. The first is to minimize adversarial perturbation (Papernot et al. 2016; Carlini and Wagner 2017; Croce and Hein 2020a), where they employ complicated approaches to allow the perturbation to be as small as possible in $L_p$ norm distances but generally have considerable computational cost. The other is to restrict the perturbation in the $\epsilon$-ball around the input (Madry et al. 2018; Dong et al. 2018; Lin et al. 2020), where they have low computational cost but introduce a large perturbation. The common point is that all approaches mentioned above adopt the first-order gradient information.

However, these approaches lack *interpretation from the point of view of DNNs*. The adversaries only looked for a more optimizable landscape to generate AE by gradients, and few of these approaches considered what features the DNN actually *learns*. It is intuitive that if the adversaries attack the features learned by the DNN, the misclassification can occur more easily. Previously, Subramanya et al. 2019 tried to use the Grad-CAM approach (Selvaraju et al. 2017) to create adversarial patches that we believe to cause a very large perturbation. Few studies have been conducted on white-box AE generation approaches with low perturbation from the perspective of interpretable approaches.

To fill this gap, in this paper we propose an interpretable and effective adversarial example generation approach, namely, the deep Taylor *D*ecomposition *I*terative white-box

*A*dversarial example *A*ttack (dubbed as DI-AA). Unlike the previous approaches, our DI-AA approach uses the interpretable saliency map through the deep Taylor decomposition approach (Montavon et al. 2017) instead of the gradient saliency map. With the guideline of the interpretable saliency map, our approach applies heuristic searches and Lagrangian relaxation of $L_p$ norm constraint to find features that should be perturbated and the adversarial perturbation is minimized, respectively.

We summarize two key contributions in DI-AA:
- We embrace the idea of finding the most contributing features by using the deep Taylor decomposition approach and then transfer it to propose an interpretable AE generation algorithm, dubbed DI-AA. As far as we know, we are the first to apply the deep Taylor decomposition interpretable approach to AE generation, which further makes the attack more transparent and interpretable.
- We propose to use the Lagrangian relaxation of $L_2$ norm so that DI-AA can constrain the perturbation. Cooperating with Lagrangian relaxation and the interpretable saliency map, DI-AA constrains the $L_2$ norm distance explicitly and the number of perturbated features, i.e., the $L_0$ norm distance, is constrained implicitly, simultaneously.

Extensive experiments are carried out to evaluate our approach. We verify DI-AA on three datasets: NSL-KDD, MNIST, and CIFAR-10, to ensure its effectiveness in a variety of scenarios. Compared to six baseline approaches, our proposed attack can attack the non-robust models with fairly low perturbation. The perturbation is closer or lower than that of the state-of-the-art AutoAttack approach (Croce and Hein 2020b). Furthermore, the proposed attack breaks the TRADES (Zhang et al. 2019) adversarial training models with the highest success rate. The generated AE can reduce the robust accuracy of robust black-box models from 16% to 31% in the black-box transfer attack.

The rest of the paper is organized as follows. Section 2 reviews the adversarial example attacks and defenses. We introduce the proposed DI-AA and evaluate it in Section 3 and 4, respectively. Section 5 concludes the paper and states future work.

## 2. Related Work

This section reviews related work on adversarial examples in attacking, defending, and understanding DNNs.

### 2.1 Adversarial Attacks

**White-box attacks.** Adversarial examples were first proposed by (Szegedy et al. 2014), who used the L-BFGS optimizer to find the perturbation. (Goodfellow, Shlens, and Szegedy 2015; Kurakin, Goodfellow, and Bengio 2017) further proposed fast and iterative AE generation approaches: Fast Gradient Sign Method (FGSM) and Basic Iterative Method (BIM), to substitute the time-consuming L-BFGS. Both methods are untargeted attacks, in which the adversary does not assign the target label and their purpose is to misclassify the DNN. Subsequently, Carlini and Wagner 2017 proposed a strong targeted attack called CW, which was the first to use the logit outputs to generate AE, and the output of AE can be specified to the target label. Since then, the white-box attacks have been separated by two goals: minimizing perturbation (Croce and Hein 2020b) and fast AE generation in $\epsilon$-ball around inputs (Madry et al. 2018; Dong et al. 2018; Lin et al. 2020). The latter is generally used in adversarial training for acquiring AE more quickly and then the adversarial training time can be efficiently reduced.

**Black-box attacks.** A mainstream of black-box attacks is the transfer attack (Dong et al. 2018; Lin et al. 2020; Wu et al. 2020). Specifically, the adversary generates AE in the local substitute DNN and then transfers AE to attack the unknown deployed models based on the AE transferability. This paper transfers our generated AE to the unknown robust models.

**AE and Interpretation.** Recently, an enormous amount of work was devoted to discovering deep networks' inner mechanisms, especially state-of-the-art convolutional neural networks (Yosinski et al. 2015; Sundararajan, Taly, and Yan 2017; Kauffmann, Müller, and Montavon 2020; Selvaraju et al. 2017). For instance, Sundararajan, Taly, and Yan 2017 focused on interpreting individual predictions. Inspired by previous work, researchers have attempted to understand AE through interpretable approaches to robust models (Boopathy et al. 2020; Dong et al. 2017).

However, few researchers paid attention to the influence of interpretable approaches on the generation of AE directly, which we think is a gap to fill. Subramanya, Pillai, and Pirsiavash 2019 used an interpretable approach, Grad-CAM (Selvaraju et al. 2017), to generate adversarial image patches. Wang et al. 2021 used Integrated Gradients (IG) (Sundararajan, Taly, and Yan 2017) to generate AE, but the IG approach has some inherent shortcomings for generating AE, which we will discuss in Section 3.3. This paper attempts to integrate interpretable approaches to generate AE while minimizing perturbation.

### 2.2 Adversarial Defenses

Adversarial training (AT) was first proposed in (Goodfellow, Shlens, and Szegedy 2015) and later improved in (Madry et al. 2018). Subsequently, several techniques were proposed to construct robust DNNs, such as ensemble training (Yang et al. 2020) and one-step training (Andriushchenko and Flammarion 2020). So far, AT, including TRADES AT (Zhang et al. 2019), is still the most effective approach to make DNNs robust (Pang et al. 2020). TRADES AT was proposed to resolve the trade-off between clean accuracy and robust accuracy. In this paper, clean accuracy denotes the accuracy of the non-robust models, and robust accuracy denotes the accuracy of the robust models. TRADES introduced a novel loss function to solve the problem that accuracy does not represent robustness (Tsipras et al. 2019). This paper uses TRADES to adversarially train our models, and we then attack the robust models with our proposed approach.

## 3. Methodology

In this section, we first describe the notation and the optimization problems in Sections 3.1 and 3.2, respectively. We then explain the motivation of integrating the deep Taylor decomposition approach and present the DI-AA approach in Sections 3.3 and 3.4, respectively.

### 3.1 Notation

In this paper, we follow the notations from (Carlini and Wagner 2017). In classification tasks, given input with $n$-dimensional features $x \in \mathbb{R}^n$ with the ground-truth label $C^*(x)$, a DNN model can be seen as a sophisticated function $F$ stacked together by multiple layers, which produces the corresponding output $y \in \mathbb{R}^m$. Here, $m$ denotes the number of classes and $\arg\max(F(x))$ is denoted as $C(x)$. Generally, the activation function of the last layer is the SoftMax function, turning the logit output to the probability distribution: $F(x) = \text{SoftMax}(Z(x))$. In our settings, we do not exploit this probability output but rather the logit output $Z(x)$ to generate AE since the SoftMax function obscures the decision boundary (Wang et al. 2018).

### 3.2 Problem Definition

Before constructing the AE, we first define the problem of finding an AE $x'$ for an input $x$. As researchers well know, AE generation can be treated as an optimization problem, where the optimization goal will vary, depending on the attacker's needs. Specifically, we want to construct the AE by maximizing the loss value of $x$ as:

$$\max \ J(F(x'), C^*(x))$$
$$\text{s.t.} \ \ C(x') \neq C^*(x); \quad (3.1)$$
$$\|x - x'\|_p \leq \epsilon$$

In this way, the loss value of $x$ can be maximized by gradient-based ascent iteration, since a larger loss value implies an incorrect classification. In this paper, we propose to use a new optimization objective described in (Wang et al. 2021) and try to optimize two things simultaneously: 1) minimize the $L_p$ norm distance, $\|x - x'\|_p$, to ensure the small perturbation; 2) minimize the logit value of $Z(x)_{C^*(x)}$ to cause a misclassification. Hence:

$$\min \ Z(x)_{C^*(x)} + c \cdot \|x - x'\|_p$$
$$\text{s.t.} \ \ x' \in \mathbb{R}^n \quad (3.2)$$

Eq. (3.2) can be viewed as a Lagrangian relaxation version of Eq. (3.1), except that Eq. (3.1) is a maximization optimization and the objective is different. Let us remark why we do not focus on the loss value but on the logit value $Z(x)$. We believe that besides increasing the loss value of $x$, there are other ways to cause a misclassification. Misclassification $C(x') \neq C^*(x)$ also occurs when the logit value $Z(x)_{C^*(x)}$ is reduced to a value which is not the maximum of $Z(x)$. The way to reduce the logit value $Z(x)_{C^*(x)}$ is more intuitive and valid than increasing the loss value. In addition, Eq. (3.2) is a suitable optimization function that can be solved by existing gradient-descent optimization algorithms such as Adam (Kingma and Ba 2015) and AdaBelief (Zhuang et al. 2020). The constraint $x' \in \mathbb{R}^n$ ensures that the AE is in the valid input domain. This definition is transparent and easy to interpret.

In Eq. (3.2), the decrease of $Z(x)_{C^*(x)}$ only achieves the untargeted attack. Here, we extend the objective function for the targeted attack, as demonstrated in Eq. (3.3).

$$\min \ c \cdot \|x - x'\|_p - Z(x)_t$$
$$\text{s.t.} \ \ x' \in \mathbb{R}^n \quad (3.3)$$

The intuition in $Z(x)_t$ of Eq. (3.3) is the opposite of Eq. (3.2). Specifically, we want $F(x)$ to output the desired target label $t$ for targeted attacks, and consequently, $Z(x)_t$ should be the maximum of $Z(x)$ to promise the output is the target label $t$. The form of Eq. (3.3) ensures that $Z(x)_t$ will increase iteratively, as $-Z(x)_t$ will decrease by the gradient descent optimizer.

In this section, we presented a broad interpretable class of problem definitions about $Z(x)$ for attacks in various settings. We concentrate on the untargeted attack form and use Eq. (3.2) as our objective instance to propose the DI-AA approach in the next section.

### 3.3 Advantages of the Deep Taylor Decomposition Approach

As mentioned in Section 2.1, Wang et al. 2021 tended to use Integrated Gradients (IG) (Sundararajan, Taly, and Yan 2017) to generate AE. However, the IG approach has some inherent weaknesses: 1) **Instability**. The IG approach needs to find its corresponding baseline data point for each dataset, e.g., a pure black image for the image dataset. Hence, the baseline dramatically impacts the performance of the IG approach. 2) **Computational cost.** The IG approach is designed to sample around the input $x$ and the frequency is user-defined. Hence, a lower frequency will affect the performance. 3) **Non-conservativeness.** The sum of contributions in each input feature by the IG approach is not equal to the model output, which we call *non-conservativeness*. Theoretically, the sum of contributions should coincide with the model output. Otherwise, it will induce unnecessary and irrelevant contributions.

To overcome these weaknesses, a more theoretically complete approach, the deep Taylor decomposition interpretation approach, was proposed in (Montavon et al. 2017). The deep Taylor decomposition approach considers the individual neuron in the DNN as a function that can be decomposed by Taylor decomposition and backward to the input. This decomposition has the following two properties:

1) Conservativeness: $\forall x: Z(x) = \sum_i R(x_i)$. Compared to the non-conservativeness mentioned above, the conservativeness property ensures that no confusing and noisy contributions are generated when the output value is backward from the last layer to the input variable. $R(\cdot)$ function refers to the contribution/relevant value of each input feature $x_i$.

2) Positive: $\forall x_i: R(x_i) \geq 0$. This property ensures that each input feature $x_i$ has a non-negative contribution or relevant value.

With *conservativeness* and *positive* properties, the deep Taylor decomposition approach has tighter restrictions than the IG approach. More importantly, compared to the IG approach, the decomposition approach does not rely on baseline input, and consequently calculates the relevant values more robustly. In what follows, we use Fig. 1 to illustrate the advantages of the decomposition approach. As shown in Fig. 1, the relevant map of the IG approach (middle picture) is not appropriate as the background should not contribute to the model outputs. In contrast, the elements in the relevant map of the decomposition approach are positive in the area around the number '2'. So, based on the distinctive advantages, we adopt this interpretable approach to generate AE.

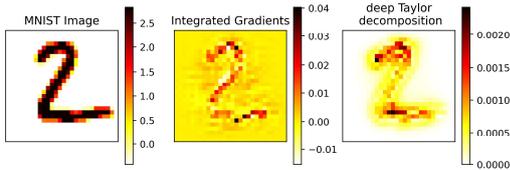

Fig. 1. Contrasting relevant maps between the IG approach and the deep Taylor decomposition in image '2' of the MNIST dataset

Since the deep Taylor decomposition approach is complex, we present it briefly here and more details can be read in (Montavon et al. 2017). Assume that the output of function $f(x)$ has been decomposed into one neuron $x_j$ of a given layer and $R_j$ is the associated relevant value. Here, $\{x_i\}$ are the neurons connected to the previous layer. We define a root point $r(x_i)$ and the root point is not specified practically. Then, the Taylor decomposition of $R_j$ is:

$$R_j = \sum_j \frac{\partial R_j}{\partial x_i}|_{r(x_i)} \cdot (x_i - r(x_i)) + \epsilon_j$$

The decomposition method spreads the above function from the output layer to the input layer, and therefore it is the deep Taylor decomposition.

### 3.4 DI-AA Approach

In this section, we propose a new AE attack approach, DI-AA. DI-AA combines the deep Taylor decomposition interpretable approach with the conventional AE generation approach. Furthermore, DI-AA reduces perturbation by constraining the $L_2$ norm distance explicitly and the $L_0$ norm distance implicitly. The algorithm flow is represented in Algorithm 1, detailed as follows.

---
**Algorithm 1: DI-AA**

**Input:** the legitimate input $x$ with the corresponding label $C^*(x)$; the model learned function $Z(\cdot)$; the perturbation rate $\varepsilon$; the iterations $T$
**Output**: adversarial example $x'$
1. $x' = \text{clone}(x)$
2. $saliency\_map = \text{DeepTaylorDecomp}(Z(x), x, C^*(x))$
3. $sorted\_map = \text{sort}(saliency\_map)$
4. $masked = \text{zeros\_like}(x)$
5. **for** $idx = 0, 1, \cdots, \text{len}(sorted\_map)$ **do**
6. $\quad masked[sorted\_map[idx]] = 1$
7. $\quad x' = \text{AEGen}(x', C^*(x), Z(\cdot), T, \varepsilon, masked)$
8. $\quad$ **if** $\text{argmax}(Z(x')) \neq C^*(x)$: **return** $x'$
9. **end for**
10. **return** $x'$

---

Algorithm 1 first initializes the original AE $x'$ by cloning the legitimate input $x$, shown in line 1. Then, in line 2, the deep Taylor decomposition approach is used to get the contributive saliency feature map, which is the standard to guide the order of the features to be perturbed. In line 3, the sort$(\cdot)$ function sorts the saliency map and returns the feature index in descending order. Here, we perturb one feature at a time in the AE generation process, which is what the iterative code in line 5 does. To guarantee the individual perturbation point, we introduce a '**0**' masked matrix $masked$ as defined on line 6. When it is required that the feature $i$ is perturbated, $masked[i] = 1$. Subsequently, a one-feature-perturbation loop is implemented to generate AE by **Algorithm 2** (line 7). An obvious advantage of perturbating one feature at a time is that in this way we can avoid pulling the redundant and unnecessary perturbation to the AE and also restrict the features to be perturbated. Therefore, the $L_0$ distance is implicitly bounded. When the AE is generated by **AEGen** (defined in **Algorithm 2**), the condition in line 8 checks if it is valid: if it can cause a misclassification, it is a suitable AE; otherwise, it will continue to iterate through the loop.

In **Algorithm 2**, **AEGen** takes $T$ iterations to generate an AE. Line 2 represents the objective function given in Eq. (3.2). Concretely, Hadamard product of one-hot encoding of $C^*(x)$ and $Z(x)$ is $Z(x)_{C^*(x)}$. Then the sum of $Z(x)_{C^*(x)}$ and the $L_2$ norm distance consists the object function. When solving the objective function, the $\text{optim}(\cdot)$ function calculates the derivatives of the objective function with regard to $x'$. The Hadamard product of the *masked* matrix and the derivatives is manipulated to obtain the perturbation. The step $\varepsilon$ plays an important role in controlling the perturbation in each iteration. The $\text{clip}(\cdot)$ function restricts each element of $x'$ within the legitimate range $[clip\_min, clip\_max]$ to satisfy the input domain. In this paper, the legitimate domain is set to $[0,1]$ in the all datasets

---
**Algorithm 2: AEGen**

**Input:** AE $x'$; the legitimate input $x$ and with the corresponding ground-truth label $C^*(x)$; the model learned function $Z(\cdot)$; the perturbation rate $\varepsilon$; the iterations $T$; the masked matrix *masked*.
**Output**: adversarial example $x'$
1. **for** $i = 1, 2, \cdots, T$ **do**
2. $\quad obj = \text{OneHot}(C^*(x)) \odot Z(x') + c \cdot L_2(x, x')$
3. $\quad grad = \text{optim}(obj, x')$
4. $\quad x' = x' - \varepsilon \cdot grad \odot masked$
5. $\quad x' = \text{clip}(x', clip\_min, clip\_max)$
6. $\quad$ **if** $\text{argmax}(Z(x')) \neq C^*(x)$: **return** $x'$
7. **end for**
8. **return** $x'$

---

## 4. Experimental Results and Analysis

This section demonstrates how to verify the effectiveness of DI-AA. We first present the datasets, models, baselines used in the experiments and hyperparameters of DI-AA. We then introduce the white-box attacks to the non-robust models and robust models. Non-robust models are models that are not adversarially trained. Robust models are the models trained by the adversarial defensive approach (TRADES in this case). Finally, to further investigate the transferability of the AE generated by DI-AA, we attack robust models in the black-box manner.

### 4.1 Setup

| Dataset | #Training | #Testing | Size | #Classes |
|---|---|---|---|---|
| NSL-KDD | 395345 | 61388 | 1×122 | 5 |
| MNIST | 60000 | 10000 | 1×28×28 | 10 |
| CIFAR-10 | 50000 | 10000 | 3×32×32 | 10 |

#Training denotes the number of the training set.

Table 1. Dataset information

**Dataset.** We focus on three datasets to validate the effectiveness of DI-AA comprehensively, as we test our approach not only on unstructured datasets like MNIST (Lecun et al. 1998) or CIFAR-10 (Alex 2009) datasets, but also on the structured dataset as NSL-KDD (Tavallaee et al. 2009). The dataset information is shown in Table 1.

**Models.** Table 2 shows the details of the KDD model and the MNIST model. As for the CIFAR-10 model, we use the standard ResNet-18 model (He et al. 2016) and omit its structure in Table 2. For detailed information about ResNet, please refer to (He et al. 2016). The AdaBelief optimizer (Zhuang et al. 2020) is implemented to boost the performance of the models in the training phase.

| Layer | KDD model | MNIST model |
|---|---|---|
| $\text{ReLU}(\text{Conv})$ | – | 3×3×32 |
| $\text{ReLU}(\text{BatchNorm}(\text{Conv}))$ | 1×4×16 | 3×3×32 |
| MaxPool | – | 2×2 |
| $\text{ReLU}(\text{Conv})$ | – | 3×3×64 |
| $\text{ReLU}(\text{BatchNorm}(\text{Conv}))$ | 1×3×32 | 3×3×64 |
| MaxPool | – | 2×2 |
| $\text{ReLU}(\text{BatchNorm}(\text{Conv}))$ | 1×3×64 | – |
| Dropout | 0.3 | 0.3 |
| $\text{ReLU}(\text{FC})$ | 1024×256 | 1024×512 |
| FC | 256×10 | 512×10 |

Table 2. The structure of the KDD and MNIST models

**Performance.** Based on the trained models, the TRADES adversarial training approach is further adopted to improve the robustness of the three models. TRADES hyperparameters are provided by the authors' default settings (Zhang et al. 2019). The results are shown in Table 3.

| Model | Clean Accuracy | Robust Accuracy |
|---|---|---|
| KDD model | 88.18% | 86.85% |
| MNIST model | 99.54% | 99.52% |
| CIFAR-10 model | 94.78% | 80.38% |

Table 3. Model Performance

**Baselines.** Comparable baselines include FGSM (Goodfellow, Shlens, and Szegedy 2015), BIM (Kurakin, Goodfellow, and Bengio 2017), PGD (Madry et al. 2018), CW (Carlini and Wagner 2017), and AutoAttack (AutoA) (Croce and Hein 2020b). The PyTorch (Paszke et al. 2019) Library implements all the approaches. FGSM, BIM, PGD, and CW approaches use the Advertorch Library (Ding, Wang, and Jin 2019). The code of AutoAttack can be found in (Croce and Hein 2020b).

**Evaluation Metrics.** To evaluate the effectiveness of DI-AA, we use three metrics: 1) *accuracy score* (ACC) to assess the transferability of AE in the black-box setting; 2) the $L_p$, $p \in \{0,1,2\}$, norm metrics to measure the perturbation; and 3) *Success Rate* (SR) to evaluate efficiency.

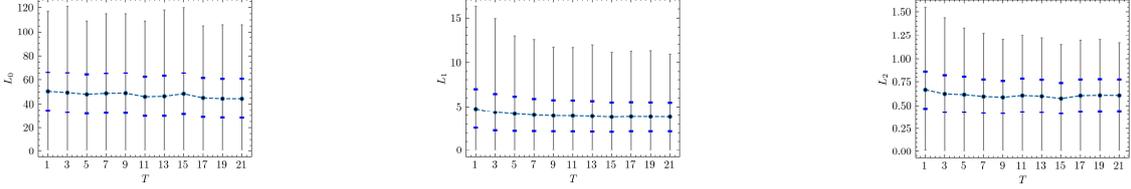

Fig. 2. $L_p$ stacked error bars of NSL-KDD dataset when iteration $T$ varies from 1 to 21

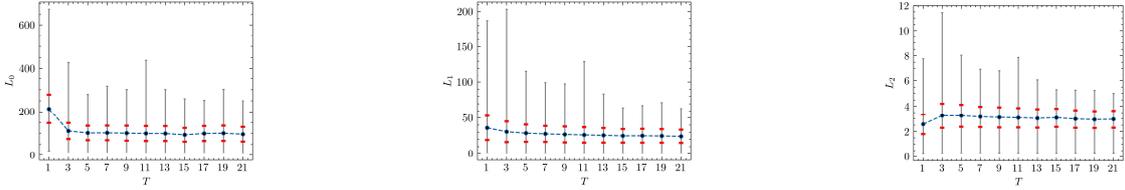

Fig. 3. $L_p$ stacked error bars of MNIST dataset when iteration $T$ varies from 1 to 21

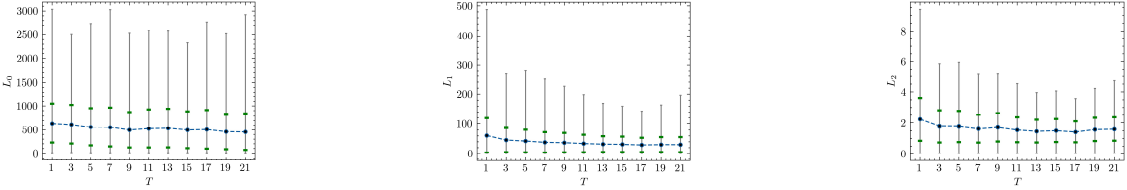

Fig. 4. $L_p$ stacked error bars of CIFAR-10 dataset when iteration $T$ varies from 1 to 21

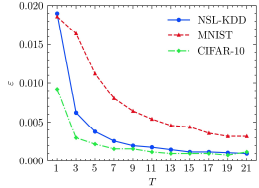

Fig. 5. The trend of $\varepsilon$ when $T$ varies from 1 to 21

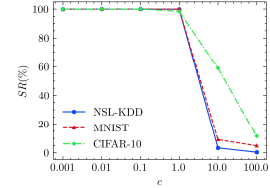

Fig. 6. The effect of $c$ to the SR metrics

| Non-robust model | $L_0$ Mean | $L_0$ Std | $L_1$ Mean | $L_1$ Std | $L_2$ Mean | $L_2$ Std | SR |
|---|---|---|---|---|---|---|---|
| Ours | 18.94 | 6.83 | **6.71** | **2.63** | **1.76** | **0.49** | **100.00%** |
| FGSM | 109.92 | 11.96 | 91.53 | 20.12 | 12.24 | 7.39 | 17.10% |
| BIM | 101.90 | 12.15 | 31.80 | 17.00 | 7.73 | 8.93 | 31.08% |
| PGD | 122.00 | 0.00 | 119.12 | 5.13 | 10.86 | 0.35 | 70.91% |
| CW | 121.58 | **3.49** | 11.40 | 10.56 | 3.60 | 6.52 | **100.00%** |
| IWA | **10.19** | 5.95 | 8.79 | 5.09 | *2.90* | 1.27 | *91.83%* |
| AutoA | 100.87 | 10.42 | 31.33 | 15.74 | 6.64 | 7.87 | 44.55% |

A. Non-robust Model

| Robust model | $L_0$ Mean | $L_0$ Std | $L_1$ Mean | $L_1$ Std | $L_2$ Mean | $L_2$ Std | SR |
|---|---|---|---|---|---|---|---|
| Ours | 26.11 | 8.96 | **9.42** | **3.44** | **2.15** | **0.55** | **100.00%** |
| FGSM | 109.67 | 11.02 | 88.44 | 21.47 | 11.59 | 6.91 | 15.84% |
| BIM | 105.29 | 10.71 | 29.91 | 16.16 | 7.06 | 8.22 | 25.40% |
| PGD | 104.62 | 7.30 | 29.41 | 14.07 | 5.76 | 6.87 | 60.28% |
| CW | 121.20 | 3.85 | 11.53 | 10.91 | 3.45 | 6.44 | **100.00%** |
| IWA | **15.76** | 9.06 | 11.32 | 6.98 | 3.08 | 1.47 | 88.54% |
| AutoA | 117.33 | **2.43** | 31.79 | 11.88 | 9.47 | 9.06 | 9.24% |

B. TRADES-robust Model

Table 4. Comparison of seven AE white-box attacks to the non-robust and robust models on the NSL-KDD dataset

| Non-robust model | $L_0$ | | $L_1$ | | $L_2$ | | SR |
|---|---|---|---|---|---|---|---|
| | Mean | Std | Mean | Std | Mean | Std | |
| Ours | 100.03 | 28.12 | *27.67* | 9.82 | *3.14* | 0.74 | **100.00%** |
| FGSM | 474.71 | 31.76 | 132.00 | 9.01 | 6.25 | 0.22 | 83.94% |
| BIM | 479.08 | 24.25 | 132.08 | 5.23 | 6.25 | 0.13 | 93.09% |
| PGD | 479.44 | 24.10 | 132.20 | *5.11* | 6.26 | **0.12** | 93.23% |
| CW | 757.24 | 19.70 | **12.18** | **4.06** | **1.13** | 0.33 | 100.00% |
| IWA | **43.96** | 43.74 | 61.27 | 57.13 | 10.72 | 5.85 | 99.83% |
| AutoA | 727.70 | **105.21** | 124.33 | 5.88 | 5.42 | 0.33 | 100.00% |

A. Non-robust Model

| Robust model | $L_0$ | | $L_1$ | | $L_2$ | | SR |
|---|---|---|---|---|---|---|---|
| | Mean | Std | Mean | Std | Mean | Std | |
| Ours | **167.02** | 68.39 | *27.52* | 13.99 | *3.58* | 1.24 | **9.08%** |
| FGSM | 445.46 | 32.07 | 121.51 | 6.53 | 5.99 | **0.16** | 1.31% |
| BIM | 483.09 | 40.41 | 133.61 | 11.15 | 6.29 | 0.27 | 1.76% |
| PGD | 483.22 | 35.02 | 133.37 | 10.31 | 6.28 | 0.25 | 1.79% |
| CW | 783.79 | **0.52** | **4.70** | **2.99** | **1.02** | 0.53 | 4.26% |
| IWA | 626.02 | 210.33 | 130.16 | 46.69 | 7.28 | 2.39 | 8.64% |
| AutoA | 646.51 | 31.13 | 140.49 | 8.69 | 6.15 | 0.28 | 6.53% |

B. TRADES-robust Model

Table 5. Comparison of seven AE white-box attacks to the non-robust and robust models on the MNIST dataset

| Non-robust model | $L_0$ | | $L_1$ | | $L_2$ | | SR |
|---|---|---|---|---|---|---|---|
| | Mean | Std | Mean | Std | Mean | Std | |
| Ours | 485.51 | 382.57 | *31.38* | 25.97 | 1.57 | 0.74 | 99.94% |
| FGSM | 3053.85 | 71.50 | 823.48 | 63.04 | 15.347 | 0.74 | 83.52% |
| BIM | 3053.37 | 71.81 | 823.27 | 63.12 | 15.346 | 0.74 | 98.44% |
| PGD | 3053.27 | 72.15 | 823.20 | 63.11 | 15.345 | 0.74 | 98.41% |
| CW | 3071.86 | **0.99** | **4.85** | 4.26 | **0.14** | 0.11 | 99.98% |
| IWA | **182.07** | 252.17 | 85.03 | 101.86 | 10.11 | 7.12 | 98.98% |
| AutoA | 3066.01 | 24.31 | 66.55 | **3.34** | 1.32 | **0.05** | **100.00%** |

A. Non-robust Model

| Robust model | $L_0$ | | $L_1$ | | $L_2$ | | SR |
|---|---|---|---|---|---|---|---|
| | Mean | Std | Mean | Std | Mean | Std | |
| Ours | 439.17 | 324.10 | 86.42 | 70.03 | 5.48 | 2.76 | **100.00%** |
| FGSM | 3060.37 | 44.64 | 858.40 | 49.19 | 15.82 | 0.59 | 85.37% |
| BIM | 3057.13 | 61.81 | 830.48 | 60.11 | 15.43 | 0.71 | 98.49% |
| PGD | 3056.44 | 65.18 | 829.40 | 61.36 | 15.42 | 0.73 | 98.53% |
| CW | 3071.88 | **0.52** | 19.94 | 13.42 | **0.89** | 0.58 | 99.99% |
| IWA | **247.02** | 354.67 | 78.53 | 105.31 | 5.84 | 5.63 | 98.52% |
| AutoA | 3060.38 | 50.06 | 332.65 | 20.61 | 6.20 | 0.25 | 98.46% |

B. TRADES-robust Model

Table 6. Comparison of seven AE white-box attacks to the non-robust and robust models on the CIFAR-10 dataset

**Hyperparameters.** Here we investigate the impact of three hyperparameters (perturbation rate $\varepsilon$, iteration $T$, and constant $c$ in the objective function) on attack capability in terms of $L_p$ norm. The first 10% test set samples of the three datasets are used to run the proposed approach on the non-robust models. When the SR score reaches 100%, we record the statistics of the $L_p$ evaluation metrics and the corresponding $\varepsilon$ and $T$. Evaluation metrics are plotted by stacked error bars to show trends, as shown in Fig. 2, 3 and 4 for the NSL-KDD, MNIST and CIFAR-10 datasets, respectively. A stacked error bar contains the four statistics: mean, standard deviation (std), minimum and maximum. These four statistics correspond to the dot in the middle, two squares around the dot, and the highest and lowest points in a vertical line, respectively. Empirically, we can see that DI-AA performs better in $17 \leq T \leq 21$, $\varepsilon = 0.001$ in the NSL-KDD dataset. In the MNIST dataset, $T = 21$ and $\varepsilon = 0.0032$ are the best choice. For the CIFAR-10 dataset, DI-AA works best when $T = 17$, $\varepsilon = 0.001$. The three curves in Fig. 6 show that our approach is effective for a constant $c$ in $(0,1]$ in all datasets.

### 4.2 White-box Attack on Non-robust and TRADES-robust Models

We now use the hyperparameters discussed in Section 4.1 to perform a white-box attack on the non-robust and the TRADES-robust models. In the NSL-KDD dataset, as shown in Table 4, our approach achieves the best SR rate with the least perturbation and outperforms the other six baselines. The poor performance of AutoAttack indicates that we should use different datasets to verify the effectiveness of the approach. In the MNIST dataset, Table 5 shows that our approach achieves the best SR score with the second-least perturbation, while CW reaches the least perturbation. As for the results of the CIFAR-10 dataset shown in Table 6, our approach to attack non-robust models achieves the thirdbest SR score and third least perturbation, which is 0.06% lower than the best SR score of AutoAttack. However, we achieve the best SR score when attacking the robust models.

In summary, when attacking non-robust and robust models, the interpretable method can help generate adversarial examples, and DI-AA can generate adversarial examples with a high success rate and relatively the least (or close to the least) perturbation.

### 4.3 Black-box attack on robust models

In this section, we transfer CIFAR-10 AE generated in Section 4.2 to attack models with defensive approaches. The results are shown in Table 7, which shows that both our approach and AutoAttack can successfully transfer AE to attack the black-box models. The results also show that our approach can decrease accuracy by about 16%~31% on robust models, generally larger than AutoAttack.

| Defensive Approach | Robust ACC | ours | AutoA | CW |
|---|---|---|---|---|
| (Ding et al. 2019) | 88.02 | **-30.54** | -30.41 | - |
| (Carmon et al. 2019) | 89.69 | **-20.01** | -16.55 | - |
| (Gowal et al. 2021) | 88.51 | **-19.36** | -12.09 | - |
| (Augustin, Meinke, and Hein 2020) | 92.41 | -15.69 | **-23.83** | - |
| (Rebuffi et al. 2021) | 89.05 | **-24.37** | -13.68 | - |
| (Zhang et al. 2020) | 89.36 | **-23.03** | -18.61 | - |
| (Zhang et al. 2019) | 84.92 | -16.94 | **-31.84** | -6.99 |

Table 7. Accuracy decrease (%) of three approaches on the black-box manner

## 5. Conclusion

In this paper, we have taken a step in investigating how to integrate the interpretable approach of deep Taylor decomposition into adversarial example generation algorithms. Extensive experiment results indicate that our proposed white-box adversarial example generation approach, DI-AA, can attack the non-robust and robust models with a high success rate and low perturbation, obtaining a perturbation closer to or lower than the previous AutoAttack approach. In addition, DI-AA can also attack the black-box models with high transfer rates. DI-AA shows that the saliency map by the interpretable approach can be the criterion to guide the generation of AE.